# Visual Transformer for Soil Classification

Aaryan Jagetia*, Umang Goenka¶, Priyadarshini Kumari‡, Mary Samuel§
*†¶Department of Information Technology, Indian Institute Of Information Technology, Lucknow, India
‡Department of Computer Science, Indian Institute Of Information Technology, Lucknow, India
§Department of Mathematics, Indian Institute Of Information Technology, Lucknow, India
*lit2019045@iiitl.ac.in, †lit2019033@iiitl.ac.in, ‡lcs2019074@iiitl.ac.in, §marysamuel@iiitl.ac.in

*Abstract*—Our food security is built on the foundation of soil. Farmers would be unable to feed us with fiber, food, and fuel if the soils were not healthy. Accurately predicting the type of soil helps in planning the usage of the soil and thus increasing productivity. This research employs state-of-the-art Visual Transformers and also compares performance with different models such as SVM, Alexnet, Resnet, and CNN. Furthermore, this study also focuses on differentiating different Visual Transformers architectures. For the classification of soil type, the dataset consists of 4 different types of soil samples such as alluvial, red, black, and clay. The Visual Transformer model outperforms other models in terms of both test and train accuracies by attaining 98.13% on training and 93.62% while testing. The performance of the Visual Transformer exceeds the performance of other models by at least 2%. Hence, the novel Visual Transformers can be used for Computer Vision tasks including Soil Classification.

## I. INTRODUCTION

In a vast number of countries, soil classification is a key concern and an emerging topic. In India, there have been various distinct types of soil. Due to industrialisation and population increase, each country's farming land is diminishing day by day, so it is important to develop methods that help in increasing productivity. Recognizing soil type might provide useful information for developing a more rational and cautious management system and applying it to cropping regions. Since soil is a mineral storehouse, it is very much useful to determine the biological, chemical, and physical properties of soil based on the kind of soil. The soil's nutrients level and pH content are the most critical factors to consider. Crop planting should be done according to soil characteristics to ensure crop success.

Farmers need to be aware of the suitable soil type for a particular crop in order to increase agricultural output, which affects the rising food demands. Soil is used by farmers to grow a broad variety of crops. Besides agricultural benefits, soil classification may aid in the civil constructed environment, since soil type influences the sort of foundation utilized when building a structure. Because the interior cannot be inspected and straightforward forecasts are not attainable, the soil is essentially unpredictable and diverse in terms of soil mechanics and geotechnical engineering. As a consequence, if an attempt is made to identify a certain type of soil without the use of a specific test, personal experience may be used when making a decision. As a result, engineers find it challenging to make accurate soil categorization decisions.

The major goal of our research is to discover the best reliable way for classifying soil using images since it has several uses in agriculture, construction and mining. There are a variety of field and laboratory methods for classifying soil, but they all have drawbacks such as time and effort. Computer-assisted soil categorization techniques are needed since they will help farmers in the field and will not take too long.

We discuss several computer-based soil classification approaches. To begin, soil classification systems based on Deep Learning and Machine Learning, such as the Convolutional neural network Model (CNN)[14], ResNet-50[12], Svm Classifiers[13], and Alexnet[11], produce cutting-edge findings. Many tests were carried out in order to discover the best architecture. Second, we used Visual Transformers(ViT)[15], a novel state-of-the-art approach for image classification. ViT is a relatively new advancement in the realm of computer vision. It is based on the Transformer Encoder Model[16], which is based on Natural Language Processing. The ViT model makes the Transformer architecture with self-attention to sequences of smaller image patches, without using convolution layers. The performance of the ViT model showed that relying on CNNs and other Machine Learning is not necessary. We also tried to find which ViT model is best suited for this particular problem of soil image classification by fine tuning the model and varying the parameters.

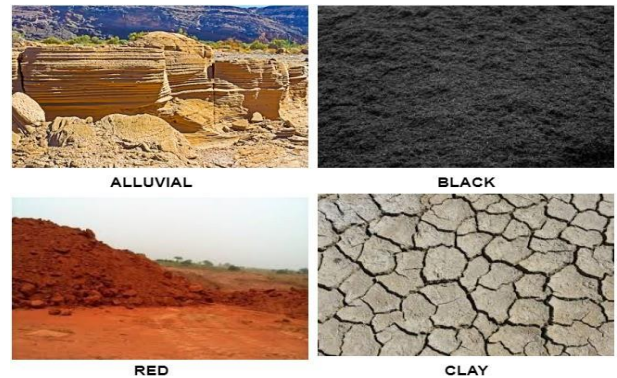

**Fig. 1:** Alluvial, Black, Red and Clay from Dataset

The following sections make up the paper: The section I explains why soil classification is vital. Details of an extensive literature survey of past studies is described in Section II. The methodologies employed in Section III explain the flow and implementation of the suggested model. In Section IV, the results and analysis of the suggested technique under various parameters are provided. The conclusions are presented in Section V.

## II. BACKGROUND AND RELATED WORK

In this section, we present the literature review, summarise, and assess the work on soil classification.

Binh Thai Pham et. al. [1] proposed a model that uses Enhanced Adaboost models. Clay content, moisture content, specific gravity, void ratio, plastic, and liquid limit parameters are used to determine the kind of soil classification. To determine soil categorization, the tree method, Adaboost, and ANN model were combined. This demonstrates that the Adaboost model's classification accuracy is well-suited to enhance the efficiency of this new technology. As a result, the Adaboost model produces the most accurate findings and can improve in the automatic classification of soil samples. The major drawback of this work is that the model is trained and tested on a relatively smaller dataset.

S. Padmavathi et. al. [2] presented work that explains how SVM may be used to identify soil kinds. The soil classifier performs picture capture, image preprocessing, extraction of features, and classification. To extract texture information from soil images, the low pass filter, Gabor filter, and colour quantization technique are utilised. The mean amplitude, HSV histogram, and standard deviation were employed as statistical measurements. However, Machine learning's potential for soil prediction would undoubtedly be increased with additional data.

Pallavi Srivastava et. al. [3] exhibited some research that gives a detailed overview of soil classifying techniques and these methods may be classified into two groups. Firstly, Image Processing and then followed by Computer Vision-based soil classification, in which image capture, segmentation, feature extraction, and soil classification are the four phases involved in these methods. Among the classification algorithms used in the approaches are random forest, maximum likelihood estimation, and k-nearest neighbour. Second, soil identification methods based on deep learning and machine learning. In terms of accuracy, the CNN model surpasses all other models, according to this research. The major challenge is that a bigger collection of soil samples is required, as Deep Learning models operate best with a large number of images.

H. K. Sharma et. al. [4] describe a model which employs SVM classification and was used to categorize soil. To examine soil they used different indicators such as HSV histogram, autocorrelation, and the wavelet movements. The suggested application provides additional features, such as soil nutrients, crop suggestions, and urea recommendations. These traits are essential for ordinary farmers since they are beneficial in farming and are simple to comprehend. The limitation comes in the form that the model works well for a defined region.

E. Theron et. al. [5] presented research which is to explore how machine vision can be utilized as a more reliable soil categorization method without relying on human error. It demonstrated that distinct types of soils are determined by the size of particles within their composition. The pipette technique and the hydrometer method were both discussed in the publication as soil categorization methods employing Stoke's Law which was compared with the machine vision system and it achieved a improving accuracy and reducing test durations by a substantial amount.

K. Chandra Mitra et. al. [6] describes a model that can forecast soil type based on land type and then suggest appropriate crops based on the prediction. A number of machine learning approaches are used to categorise soils, including weighted k-Nearest Neighbor (k-NN), Bagged Trees, and Gaussian kernel based Support Vector Machines (SVM). The suggested SVM-based technique outperforms numerous mentioned methods. In order to make this model more dependable and precise, data from additional districts should be incorporated.

Noor Akhmad Setiawan et. al. [7] examined many machine learning approaches for identifying soil types in a study. For this category, support vector machine (SVM), neural network, decision tree, and naive bayesian techniques are described and assessed. The result indicated that SVM beats the other methods when using a linear function kernel with achieving accuracy score of 82.35%. SVM accuracy did not increase considerably with attribute selection, and neither did class reduction. Furthermore, there is an option to improve Kernel Function in this study.

Dr. P K Arunesh et. al. [8] discussed three data mining classification methods, including Naive Bayes, JRip, and J48, in a paper. These algorithms are used to extract information from soil data and look at two different types of soil: red and black. The outcome for the model is that the JRip model can offer more trustworthy data outputs, and the forecast's Kappa Statistics have been raised. The limitation comes to be that the author used a dataset that consists of only two classes which is seems unrealistic.

Shina Inazumi et.al. [9] presented work which focuses on using neural networks for soil classification using image recognition. 1000 photos and three types of soil were included in the dataset. The paper implemented a neural network and varied the image size during learning. The paper lacks in varying the model by adding or removing convolution and pooling layers. Furthermore, the article did not make any adjustments to factors that may alter the model's accuracy.

Aditya Motwani et. al. [10] proposed a model which worked on the same dataset mentioned in the paper below. There are four various kinds of soil in the dataset. The work initially used a baseline CNN model and then improved the architecture to achieve an accuracy of The Symmetrical CNN architectures perform well, with a 95.21% success rate. Also, the paper compared their results on various other architectures such as SVM and AlexNet. However, the paper did not mention data augmentation and how the model dealt with overfitting.

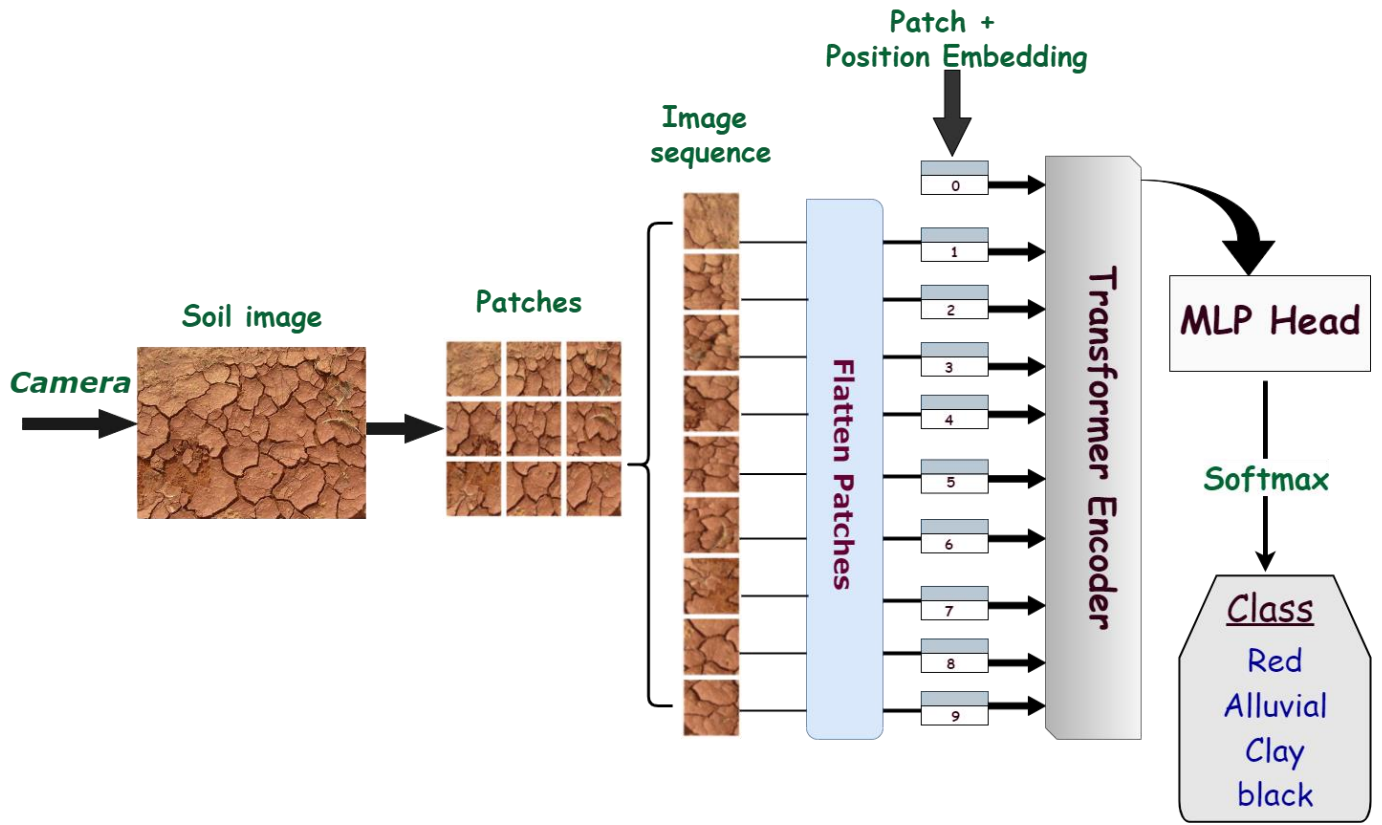

**Fig. 2:** The pipeline of the system. Image captured from camera is fed into the model. Image is converted into patches. Patches are then given input in the form of sequence after which they are flattened. To the sequence of vectors, positional embedding is added and then passed to Transformer Encoder followed by a Classifier.

### III. PROPOSED WORK

This section describes the working of Visual Transformer as mentioned in Fig. 2 and gives a brief overview about other models such as SVM, CNN, ResNet and AlexNet. The section also gives a brief description about preprocessing of dataset, structure of Visual Transformer and parameters that can affect the accuracy of the model.

*A. Dataset*

We are making use of the Soil Dataset[17] in this procedure, which contains photos of four different types of soil, As indicated in Fig. 1, alluvial soil, red soil, black soil, and clay soil are all types of soil. It contains around 1000 photos and is a balanced dataset, meaning that each class has roughly equal numbers of samples. This collection comprises annotated examples, which means that each image has a label associated with it.

*B. Working of Visual Transformer*

The functioning architecture of a Visual Transformer is based on the working nature of the transformer used in natural language processing. It turns out that if applied directly to a sequence of picture patches, a pure transformer could do extremely well on image classification tasks. To implement a Visual Transformer for image classification, loading the Soil Dataset is the first and most important step. After extracting and analyzing the dataset, because there is a limited amount of training data, we employ data augmentation to increase the amount, quality, and relevance of the data. It facilitates the model's capacity to generalise to new data. We'll resize, rotate, flip, and normalize the photos at this step. Furthermore, hyperparameters such as learning rate, weight decay, batch size, number of epochs, patch size, number of heads, transformer layers, and others will be defined. For instance, we set the picture size to 72 and the projection size to 64.

Then, following the preprocessing, we will create a network using an MLP network and a layer that will divide our photos into patches. The input picture must be partitioned into patches of the same shape using the Patch Maker in vision transformer. Patches are utilised for input representation [18]. The patches are then flattened using a linear transformation matrix. In addition, By projecting a patch onto a vector of size projection dimension, the PatchEncoder layer will linearly convert it. From the flattened patches, lower-dimensional linear embeddings was created. It also adds a learnable position embedding to the projected vector. We employ the conventional technique of adding an extra learnable classification token to the sequence in order to conduct classification.

The sequence is however given to the Transformer Encoder, which has multi-head self-attention layers and dense layers as well as a series of vectors as outputs. The architecture of Transformer Encoder is mentioned in Fig. 3. Aside from the layers, transformers improve performance by using skip connections and normalization. This is the vector outputsequence of the Transformer Encoder network. We just need one vector, the feature vector, which is fed through the softmax classifier, which outputs a vector with a size of 4, which is also the number of classes. We tested this architecture on different variants of Visual Transformers by varying the number of heads, patch size and transformer layers. Fig. 3 depicts the layer diagram for the following ViT model with transformer layers = 8.

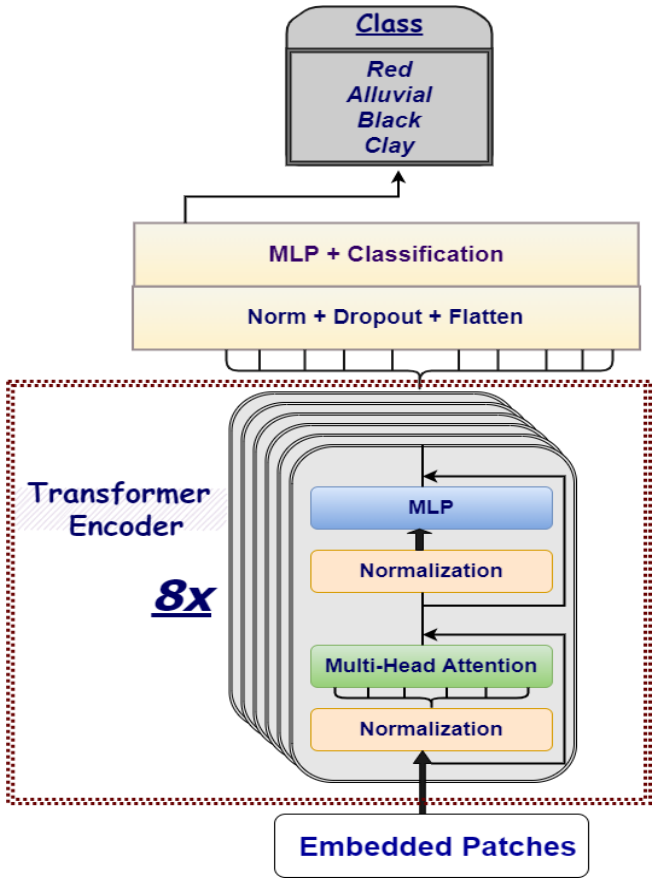

**Fig. 3:** Visual Transformer architecture for 8 Transformer Layers

C. *Other models:*

1) AlexNet[11] uses CNN. While using this model we first need to resize the image and change the number of output classes, which is 4 in this case as compared to 1000 classes in the original model. Fig. 4 mentions the structure of AleNet model.
2) Residual Neural Network(ResNet-50) [12] is a CNN. In order to reduce error rate, ResNet-50 skips connections, or short path to avoid some layers. Before ResNet-50, in order to reduce error more layers used to be added. Most of the ResNet-50 are implemented with 2 or 3- layer avoidance that consists of non-linearities (ReLU) and normalization in middle. Fig. 4 shows the model architecture of ResNet-50.
3) Support Vector Machine: SVM[13] is a Supervised Machine Learning Algorithm. Initially the images were reshaped into 200*200*3 dimensions. The soil images were converted into numerical arrays of 12000*1 dimension. The features are then sent into the SVM model as input.
4) Convolutional Neural Networks: For the CNN[14] model, the images were first converted into 300*300*3 size. After that, the images are converted into numerical arrays, which are input into the CNN model. The Adam optimization algorithm is used to train the model with learning rate for all architectures as 10Ë-4 and no. of epochs as 25. The model has 9 layers. The model architecture of the 3 models i.e. AlexNet, CNN and ResNet-50 are mentioned in Fig. 4.

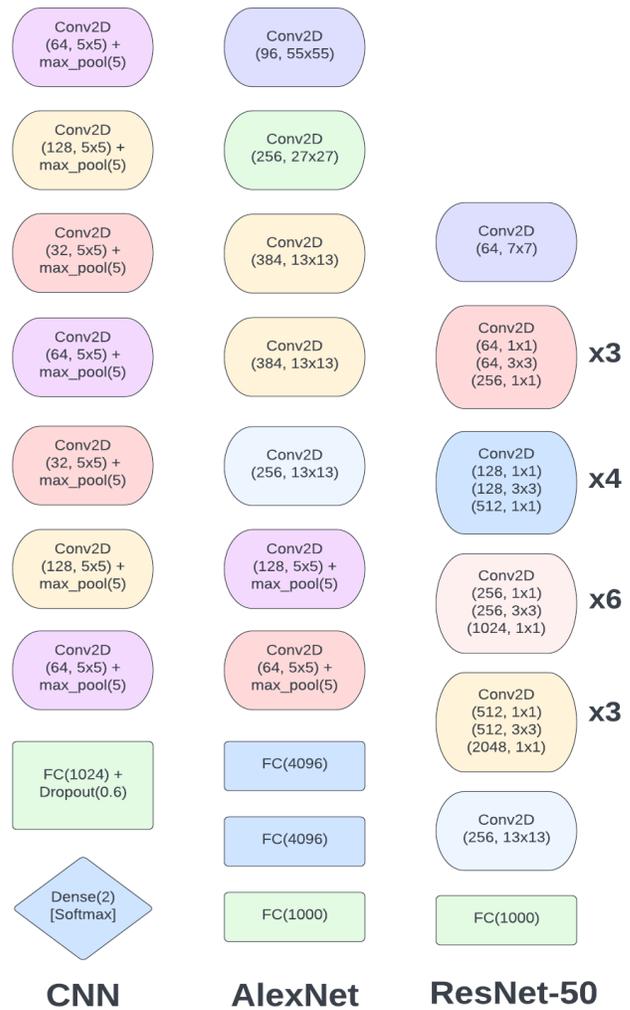

**Fig. 4:** Layer Diagram of CNN, AlexNet and ResNet-50

## IV. EXPERIMENT

This section describes the results achieved using various modifications of Visual Transformer. The results of the Visual Transformer are also compared with AlexNet, SVM, ResNet-50 and, CNN.

### A. Results on different Visual Transformers(ViT):

3 types of Visual Transformer models are constructed:
- ViT-8(Visual Transformer with patch size=8).
- ViT-12(Visual Transformer with patch size=12).
- ViT-16(Visual Transformer with patch size=16).

The results when the number of Transformer Layers=8 are mentioned in Table I and Transformer Layers = 12 are mentioned in Table II. The combined graph of all the models is plotted in Fig. 5. We have used sparse categorical accuracy metric. From the graph it is clearly visible that ViT-8, ViT-12 and ViT-16, all show an accuracy of above 96% with the exception of some outliers. ViT outperforms all other models in terms of accuracy. The best accuracy of 98.13% was achieved by ViT-8 with 8 Transformer Layers and 2 Heads. Overall ViT-8 performed better than ViT-12 and ViT-16. Among ViT-12 and ViT-16, ViT-12 performed better.

For comparison we have varied 3 parameters
- Patch Size.
- Number of Heads.
- Transformer Layers.

**TABLE I:** Accuracy scores for Transformer Layer = 8

| Heads | ViT-8 train | Vit-8-Test | Vit-12 Train | Vit-12 Test | Vit-16 Train | Vit-16 Test |
|---|---|---|---|---|---|---|
| 2 | 98.13% | 93.62% | 97.2% | 89.89% | 97.2% | 87.77% |
| 4 | 98.09% | 89.36% | 97.82% | 87.23% | 96.58% | 85.64% |
| 8 | 96.73% | 89.36% | 97.67% | 89.89% | 97.51% | 86.7% |

**TABLE II:** Accuracy scores for Transformer Layer = 12

| Heads | ViT-8 train | Vit-8-Test | Vit-12 Train | Vit-12 Test | Vit-16 Train | Vit-16 Test |
|---|---|---|---|---|---|---|
| 2 | 98.13% | 89.36% | 98.09% | 92.02% | 96.73% | 84.04% |
| 4 | 97.98% | 89.89% | 96.73% | 88.83% | 96.27% | 84.57% |
| 8 | 98.13% | 92.02% | 95.33% | 86.17% | 96.73% | 87.23% |

### B. Comparison with other models:

We have compared ViT-8, ViT-12 and ViT-16 with AlexNet, ResNet, SVM and, CNN as shown in Table III. AlexNet showed an accuracy of only 31.38%. SVM fared better but achieved an overall accuracy of 87.12% as compared to 92.21% achieved by ResNet-50. Other than ViT, CNN was the most accurate model, with an accuracy of 94.78%. The problem with CNN was that it was overfitting and showing a huge variation in test accuracy, which is 86.45% in this case. This comparison proved that ViT-8, ViT-12 and ViT-16 performed better than other models both in terms of training and testing.

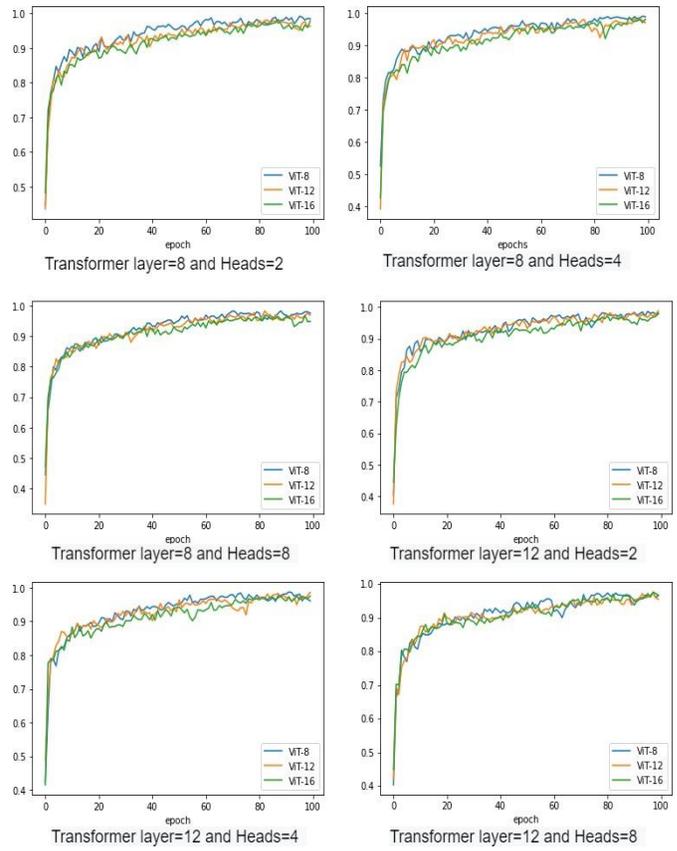

**Fig. 5:** Accuracy Plot of all the ViT Models.

**TABLE III:** Comparison of Accuracy scores of other models with Visual Transformer

| Model | Train Accuracy | Test Accuracy |
|---|---|---|
| AlexNet | 31.38% | 30.21% |
| SVM | 87.12% | 83.48% |
| ResNet-50 | 92.21% | 86.98% |
| CNN | 94.78% | 86.45% |
| ViT-8 | 98.13% | 93.62% |
| ViT-12 | 98.09% | 92.02% |
| ViT-16 | 97.51% | 86.70% |

## V. CONCLUSION AND FUTURE SCOPE

We employed a Visual Transformer for image classification, which outperforms SVM, CNN, ResNet-50, and, other Machine Learning models in terms of accuracy. The model had a precision of 98.13%, which was 2% higher than any other models. The majority of the models, with the exception of Visual Transformer, are based on Convolutional Neural Networks. The formulation of an image classification problem as a sequential problem utilizing image patches as tokens and processing it by a Transformer is the major engineering component of this study. If we introduce a lot of noise to the data, for example, if it's raining outside and a farmer wishes to categorize soil at night our model can fail to classify accurately.